\documentclass[fleqn,11pt]{wlscirep}
\usepackage[utf8]{inputenc}
\usepackage[T1]{fontenc}
\usepackage{graphicx}
\usepackage{cite}
\usepackage{amsmath,amssymb,amsfonts}
\usepackage{algorithmic}
\usepackage{graphicx}
\usepackage{textcomp}
\usepackage{booktabs}
\usepackage{url}
\usepackage[colorinlistoftodos]{todonotes}
\usepackage{bm}
\usepackage{stmaryrd} 
\usepackage{multirow}
\usepackage{booktabs}
\usepackage{graphicx}
\usepackage[flushleft]{threeparttable}
\usepackage{mwe}
\usepackage{xr}

\title{Detecting multi-timescale consumption patterns from receipt data: A non-negative tensor factorization approach}

\author[1]{Akira Matsui}
\affil[1]{Department of Computer Science, University of Southern California, Los Angeles, CA, USA}

\author[2,*]{Teruyoshi Kobayashi}
\affil[2]{Department of Economics, Center for Computational Social Science, Kobe University, Kobe, Japan}
\author[3]{Daisuke Moriwaki}
\affil[3]{AI Lab, CyberAgent, Inc., Shibuya, Tokyo, Japan.}

\author[1,4,5]{Emilio Ferrara}
\affil[4]{Information Sciences Institute, University of Southern California, Los Angeles, CA, USA.}
\affil[5]{Department of Communication, University of Southern California, Los Angeles, CA, USA.}

\affil[*]{Corresponding author: kobayashi@econ.kobe-u.ac.jp}

\begin{abstract}
Understanding consumer behavior is an important task, not only for developing marketing strategies but also for the management of economic policies. Detecting consumption patterns, however, is a high-dimensional problem in which various factors that would affect consumers' behavior need to be considered, such as consumers' demographics, circadian rhythm, seasonal cycles, etc. Here, we develop a method to extract multi-timescale expenditure patterns of consumers from a large dataset of scanned receipts. We use a non-negative tensor factorization (NTF) to detect intra- and inter-week consumption patterns at one time. The proposed method allows us to characterize consumers based on their consumption patterns that are correlated over different timescales.
\keywords{Multi-timescale patterns \and consumer behavior \and consumption expenditure \and non-negative tensor factorization}
\end{abstract}

\begin{document}

\flushbottom
\maketitle

\thispagestyle{empty}

\section*{Introduction}

 Consumption has been extensively studied in multiple research disciplines, and their viewpoints differ from one another.
 Macroeconomists, for example, consider that individual consumers' decision determines the economic condition at the macroscopic level~\cite{mankiw2003macroeconomics}. In marketing studies, on the other hand, analyzing the shopping behavior of individual consumers is essential to gain insight into business strategy~\cite{bell1998shopping}. Researchers also study consumption at different time scales; economists often assume that representative individuals live infinitely long to investigate life-long consumption paths, while business researchers are interested in shorter practical time scales. 
 
 Many studies point out that consumption patterns change in accordance with the consumer's stage of life~\cite{attanasio2010consumption, hurd2013heterogeneity, aguila2011changes}. Arguably, young people having a child would go to supermarkets more frequently than elderly people. Income level of an individual would also affect how often and how much they spend for what. Different demographic characteristics may therefore exhibit different dynamical patterns of expenditure, and this leads us to conjecture that we could infer consumers' demographic properties from their dynamical expenditure patterns.

To understand the consumption behavior of individuals with different demographic properties, we explore the following  research questions:

\begin{description}
   \item[\textbf{RQ1}:] Does consumers' expenditure behavior exhibit dynamical patterns over multiple timescales?
   \item[\textbf{RQ2}:] Do the dynamical patterns reflect demographic differences?
   \item[\textbf{RQ3}:] What demographic factors characterize the expenditure patterns?
\end{description}

To answer these research questions, we develop a non-negative tensor factorization (NTF) method to detect multi-timescale patterns of consumers' expenditure at intra- and inter-week scales. We employ the PARAFAC decomposition as a means to factorize a three-way tensor representing the actual expenditure data~\cite{bro2003new,kolda2009tensor,lim2009nonnegative}. The NTF method has been widely used to mine temporal patterns in different social contexts, such as face-to-face contacts among humans~\cite{gauvin2014detecting, sapienza2018estimating}, online communications~\cite{panisson2014mining}, online game~\cite{sapienza2018NTF} and students' life in a university~\cite{hosseinmardi2019discovering}. However, mining multi-timescale patterns has not been done so far, except for the study uncovering the intra- and inter-day transaction patterns of banks~\cite{kobayashi2018extracting}.
 
 In our model, the $(i,j,k)$-th element of a tensor corresponds to the number of items purchased by consumer $i$ on $j$th day of week $k$. The NTF allows us to know how the intra-week expenditure behavior is associated with the inter-week patterns and how many such multi-timescale patterns exist. We argue that different multi-timescale patterns may come from different demographic characteristics of consumers, such as gender, marital status, and age. This suggests that people in different stages of life indeed spend differently both at intra- and inter-week scales.  

\section*{Related Work} 

Maximizing aggregate consumption is a primary goal for policymakers and is considered to contribute to social welfare~\cite{woodford2011interest,walsh2017monetary}.
 Economists often model consumer behavior as a solution to a utility maximization problem with infinite horizon~\cite{campbell1989consumption, johnson2006household, hsieh2003consumers,walsh2017monetary}. Using a formal framework based on a utility maximization problem, economists have been discussing how consumers form and follow consumption habits~\cite{alvarez2004habit,havranek2017habit}, including whether or not such an explicit dynamical pattern exists\cite{dynan2000habit,guariglia2002consumption, carrasco2005consumption, browning2007habits, crawford2010habits, havranek2017habit}. Various studies also point out that consumption patterns tend to change according to the consumer's stage of life~\cite{attanasio2010consumption, hurd2013heterogeneity, aguila2011changes}.

Marketing scientists study consumer behavior from a more business-oriented viewpoint. For instance, they model the expenditure pattern of targeted consumers to predict the effect of a business strategy, such as a recommendation system, on actual consumption~\cite{fong2011web}. Models of consumer behavior in marketing studies incorporate various factors, including the structure of consumers' network~\cite{rosenquist2010spread, bressan2016limits}, self-revealed information in social media~\cite{de2016characterizing, silva2017large}, and spatial information regarding the consumer's geographical location~\cite{wagner2014spatial}. Among many factors that could explain the observed consumption patterns, the sequence of temporal actions has been particularly studied to understand consumers? dynamic behavior~\cite{moe2003buying, moe2004capturing,olbrich2011modeling, senecal2005consumers ,benson2016modeling}. A dynamical model has also been used to predict consumers' future activity~\cite{platzer2016ticking}. Notably, some studies point out that there are temporal patterns of shopping activity at the intra-week scale, i.e., day-of-week effects~\cite{kahn1989shopping, namin2019hidden, bogomolova2016socio}. 

In this study, we employ a non-negative tensor factorization (NTF) method~\cite{kolda2009tensor,lim2009nonnegative} to uncover hidden patterns in our receipt data. We represent consumers' expenditure data as a 3-way tensor, which will be detailed in the following section. NTF is widely used to mine temporal patterns in face-to-face contacts~\cite{gauvin2014detecting, sapienza2018estimating}, financial transactions~\cite{kobayashi2018extracting}, online communications~\cite{panisson2014mining} and online games~\cite{sapienza2018NTF}. Based on the decomposed patterns from our consumption data, we show that consumers with different demographics have different consumption patterns.

\section*{Data}
Our dataset is constructed from the receipt data scanned through a bookkeeping smartphone application \texttt{Dr.Wallet}~\cite{walletHP}. This application allows users to digitize the record of their purchases by scanning receipts using smartphones or tablet PCs. 
Item names listed in receipts are annotated and documented by human workers. The dataset contains the prices, the name of each item and the date when the receipt has been scanned. 
There are in total 2,796,008 purchased items recorded by 2,624 users from April 1, 2017 to January 21, 2018. 
The data also contains the demographic attributes of the users such as gender, marital status and age range. Table~\ref{tab:basic_stat} shows the basic statistics and the demography of users.

\begin{table}[]
\centering
    \caption{Basic statistics of receipt data collected from Dr. Wallet between April 1, 2017 and January 21, 2018. Total number of purchased items is 2,796,008. Age range is in ascending order, i.e., 1 and 6 denote the youngest and the oldest cohorts, respectively.} 
    \label{tab:basic_stat}
\begin{tabular}{@{}lrrrcr@{}}
\toprule
\multicolumn{1}{c}{}            & Category     & \#users &                            & Cohort & \#users \\ \midrule
\multirow{1}{*}{Gender}         & Female        & 1,887   & Age             & 1        & 69     \\
                                & Male          & 737    &                            &2       & 690   \\
\multirow{1}{*}{Marital status} & Married      & 1,628   &                            & 3        & 824    \\
                                & Unmarried       & 996     &                            & 4        & 673    \\
\multirow{1}{*}{Child}          & No children     & 1,345    &                            & 5        & 331     \\
                                & With children & 1,279    &                            & 6        & 137     \\
                                                  \\ 
Total \#users                           &              &2,624    &                            & 
                                &         \\ \bottomrule
\end{tabular}

\end{table}

\section*{Methods}

\subsection{Tensor representation of consumption expenditure}

\begin{figure}  
 \includegraphics[width=\linewidth]{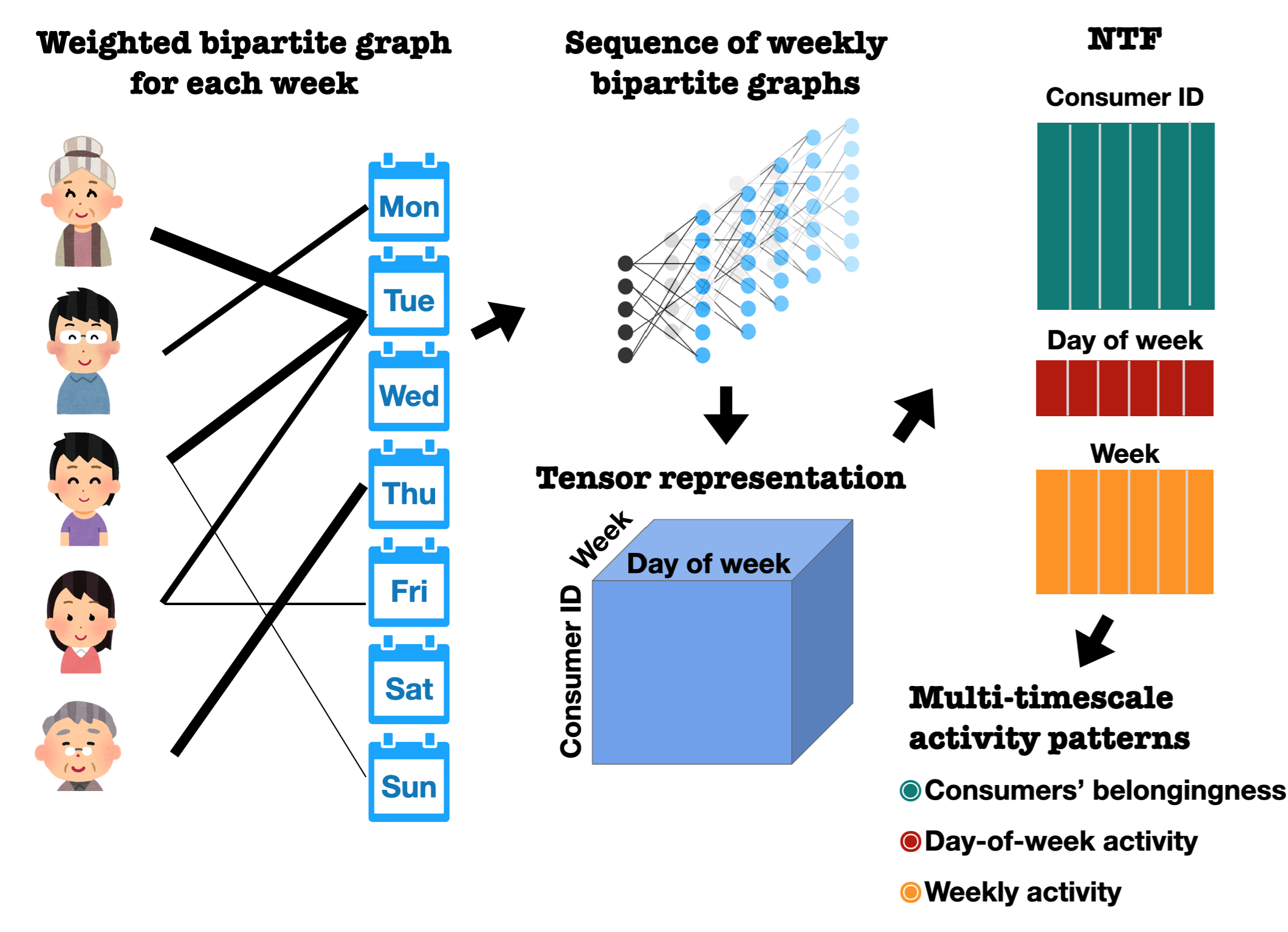}
 \caption{Schematic of NTF for extracting intra- and inter-week expenditure patterns.} 
  \label{fig:ntf_sketch}
\end{figure}

Our study aims to detect dynamical patterns from our shopping record dataset. To pursue this goal, we use a non-negative tensor factorization (NTF) to obtain the latent factors that would reflect the characteristic expenditure patterns across different attributes of consumers~\cite{kolda2009tensor,lim2009nonnegative,gauvin2014detecting,sapienza2018NTF}.
Here, we try to extract multi-timescale patterns that would exist at intra- and inter-week scales~\cite{kobayashi2018extracting}. We represent the users' shopping records by a 3-way tensor, whose size is given by $I\times J\times K$, where $I=$\#consumers ($=2,624$), $J=$\#days in a week ($=7$) and  $K=$\#weeks ($=42$).  The constructed 3-way tensor is interpreted as representing a sequence of weekly bipartite networks in each of which the nodes denoting the days of the week are connected to users with edge weights being the number of purchased items (Fig.~\ref{fig:ntf_sketch}).

\subsection{Non-negative tensor factorization}

 The NTF method decomposes tensor $\mathcal{X}\in \mathbb{R}_{+}^{I \times J \times K}$ into latent factors that characterize the activity patterns of the corresponding mode. Each element of the tensor is denoted by $x_{i j k} \in \mathcal{X}$. In our model, $x_{ijk}$ denotes the number of items purchased by user $i$ on $j$-th day of week $k$. We employ the PARAFAC decomposition as an NTF algorithm throughout the analysis~\cite{bro2003new,kolda2009tensor}. The PARAFAC decomposition is an approximation method that expresses $\mathcal{X}$ as a sum of rank-one non-negative tensors $\{\hat{\mathcal{X}_{r}}\}_{r=1}^{R}$:
 \begin{equation}
    \mathcal{X} \approx \sum_{r=1}^{R} \hat{\mathcal{X}_{r}} = \sum_{r=1}^{R} \mathbf{a}_{r} \circ \mathbf{b}_{r} \circ \mathbf{c}_{r},
    \label{eq:parafac}
\end{equation}

where $R$ denotes the number of components, and 
$\mathbf{a}_{r}\in\mathbb{R}_{+}^{I \times 1}$, $\mathbf{b}_{r}\in\mathbb{R}_{+}^{J \times 1}$ and $\mathbf{c}_{r}\in\mathbb{R}_{+}^{K \times 1}$ represent the $r$-th component factors that respectively encode the membership of a user to a component, intra- and inter-week activity levels. The operator $\circ$ represents outer product.

Let $\mathbf{A}\in \mathbb{R}_{+}^{I \times R}$, $\mathbf{B}\in \mathbb{R}_{+}^{J \times R}$ and $\mathbf {C}\in \mathbb{R}_{+}^{K \times R}$ be the factor matrices, whose $r$-th columns are vectors $\mathbf{a}_{r}$, $\mathbf{b}_{r}$ and $\mathbf{c}_{r}$, respectively. The factor matrices $\mathbf{A}$, $\mathbf{B}$ and $\mathbf{C}$ are obtained by solving the following minimization problem with non-negativity constraints:
\begin{equation}
 \min_{\mathbf{A}\geq 0,\mathbf{B}\geq 0, \mathbf{C}\geq 0} \lVert \mathcal{X} - \llbracket \mathbf{A},\mathbf{B},\mathbf{C}\rrbracket \rVert_{\rm F}^{2},
\end{equation}
where $\|\cdot\|_{\rm F}$ denotes the Frobenius norm, and $\llbracket \mathbf{A},\mathbf{B},\mathbf{C}\rrbracket$ represents the Kruscal form of the tensor decomposition (i.e., the right-hand side of Eq.~\ref{eq:parafac}). To solve this problem, we use the alternating non-negative least squares (ANLS) with the block principal pivoting (BPP)~\cite{kim2012fast}.

\subsection{Number of components}

We utilize the Core-Consistency Diagnostic to determine an appropriate number of components, $R$~\cite{bro2003new}. 
The basic idea of the Core-Consistency measure is to quantify the difference between PARAFAC decomposition and a more general decomposition, namely the Tucker3 decomposition~\cite{bro2003new}. The Tucker3 decomposition is more flexible than PARAFAC because it allows for correlations between different components. If PARAFAC and Tucker3 return similar decomposition, then the PARAFAC model is considered to be a good approximation of the original tensor (i.e., ignoring correlations among components would be justified).

 For the PARAFAC decomposition, the $(i,j,k)$ element of the tensor can be written as
\begin{equation}\label{eq:cc_parafac}
x_{i j k}=\sum_{n=1}^{R} \sum_{m=1}^{R} \sum_{p=1}^{R} \lambda_{n m p} a_{i n} b_{j m} c_{k p},
\end{equation}
where $\lambda_{n m p}$ denotes a product of Kronecker delta, i.e., $\lambda_{n m p}=\delta_{n m} \delta_{m p} \delta_{n p}$, where $\delta_{nm}$ is the Kronecker delta that takes one if $n=m$, and $0$ otherwise.
Note that $\lambda_{n m p}$ takes 1 if $n=m=p$ and $0$ otherwise, so $\lambda_{nmp}$ is the $(n,m,p)$ element of the superdiagonal binary tensor $\mathcal{L}$. 

For the Tucker3 model, the $(i,j,k)$ element of the tensor is generally written as
\begin{equation}\label{eq:cc_tucker3}
x_{i j k}=\sum_{n=1}^{R_{n}} \sum_{m=1}^{R_{m}} \sum_{p=1}^{R_{p}} g_{n m p} a_{i n} b_{j m} c_{k p},
\end{equation}
where $g_{nmp}$ may not be expressed by a product of the Kronecker delta. $g_{nmp}$ is an element of the core tensor ${\cal G}$ obtained by the Tucker3 algorithm~\cite{kolda2009tensor}.

The Core-Consistency (CC) quantifies the difference between PARAFAC and Tucker3 decomposition by computing the distance between ${\cal L}$ and ${\cal G}$ as 
\begin{equation}
\mathrm{CC}=100 \times\left(1-\frac{\sum_{n=1}^{R} \sum_{m=1}^{R} \sum_{p=1}^{R}\left(g_{n m p}-\lambda_{n m p}\right)^{2}}{R}\right).
\label{eq:cc}
\end{equation}
Note that the number of components $R$ is common for all modes in both the PARAFAC and the Tucker3 decomposition, i.e., $R_{n}=R_{m}=R_{p}=R$. If the PARAFAC and the Tucker3 methods yield exactly the same decomposition, then ${\rm CC}=100$~\cite{bro2003new}. In general, CC value decreases with $R$ because interactions between components tend to be more evident as the number of components increases.

\begin{figure}[tbh] 
 \centering
  \includegraphics[width=9cm]{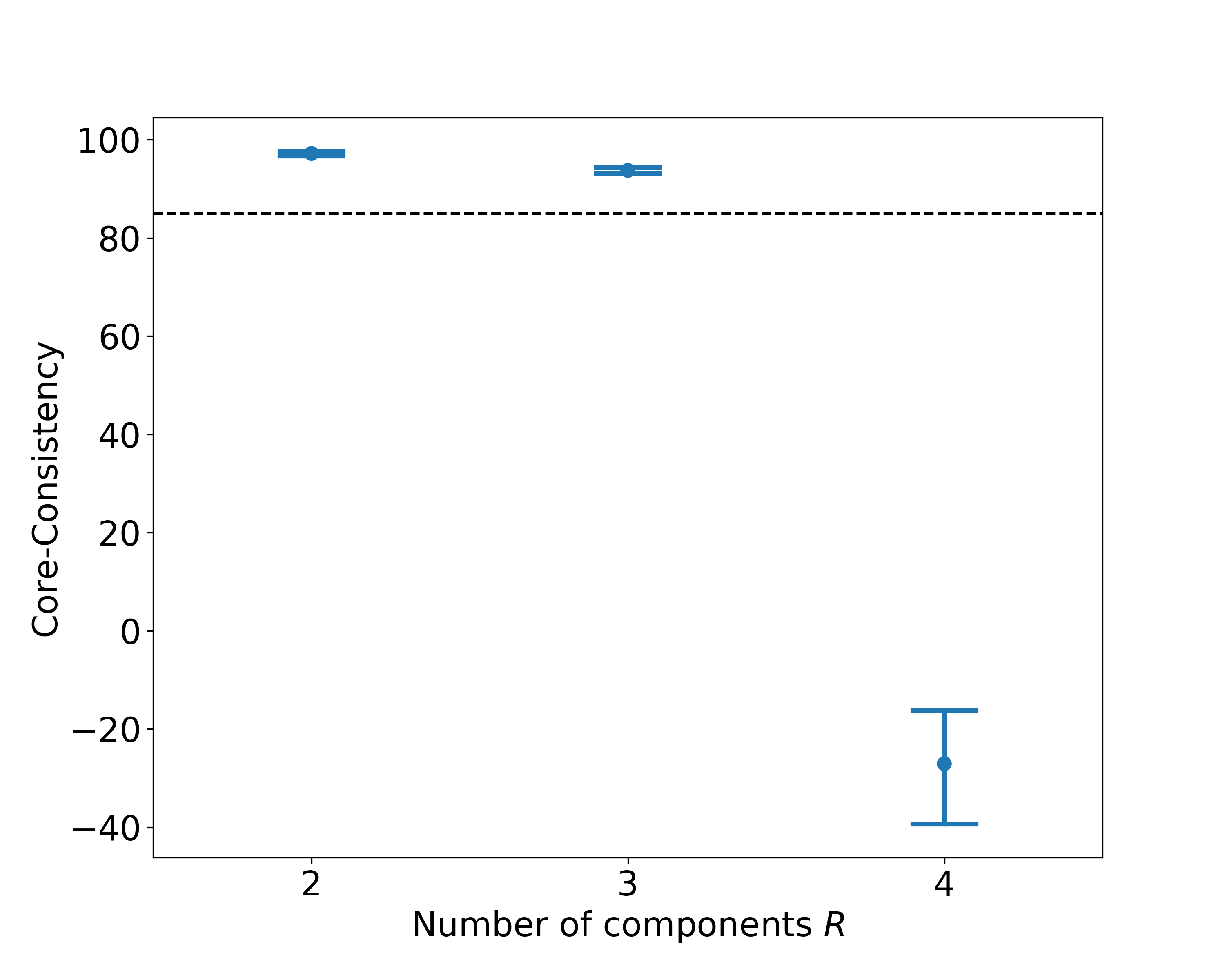}
  \caption{Core-consistency averaged over 20 runs of PARAFAC decomposition. Error bar denotes 95\% confidence interval. Horizontal dashed line denotes ${\rm CC}=85$. }
  \label{fig:cc_selection}
\end{figure}

\section*{Results}

\subsection{Core-Consistency}

 The CC values for our NTF results with different rank size $R$ are shown in Fig.~\ref{fig:cc_selection}. Since the solution for the PARAFAC decomposition is not unique due to randomly selected seeds, we run the decomposition algorithm 20 times for each $R$ and calculate the mean of the CC value with the 95\% confidence interval. The result indicates that $R=3$ would be the best choice because the CC value is larger than a rule-of-thumb threshold ($=85$)~\cite{sapienza2018estimating} up to $R=3$ and turns negative for $R=4$. Therefore, we set $R=3$ in the following analysis. We have repeated this procedure multiple times and confirmed that the results presented in the rest of the paper is qualitatively unaffected by the randomness of seeds.

\subsection{Multi-timescale expenditure patterns}

\begin{figure}[tbh] 
 \includegraphics[width=\linewidth]{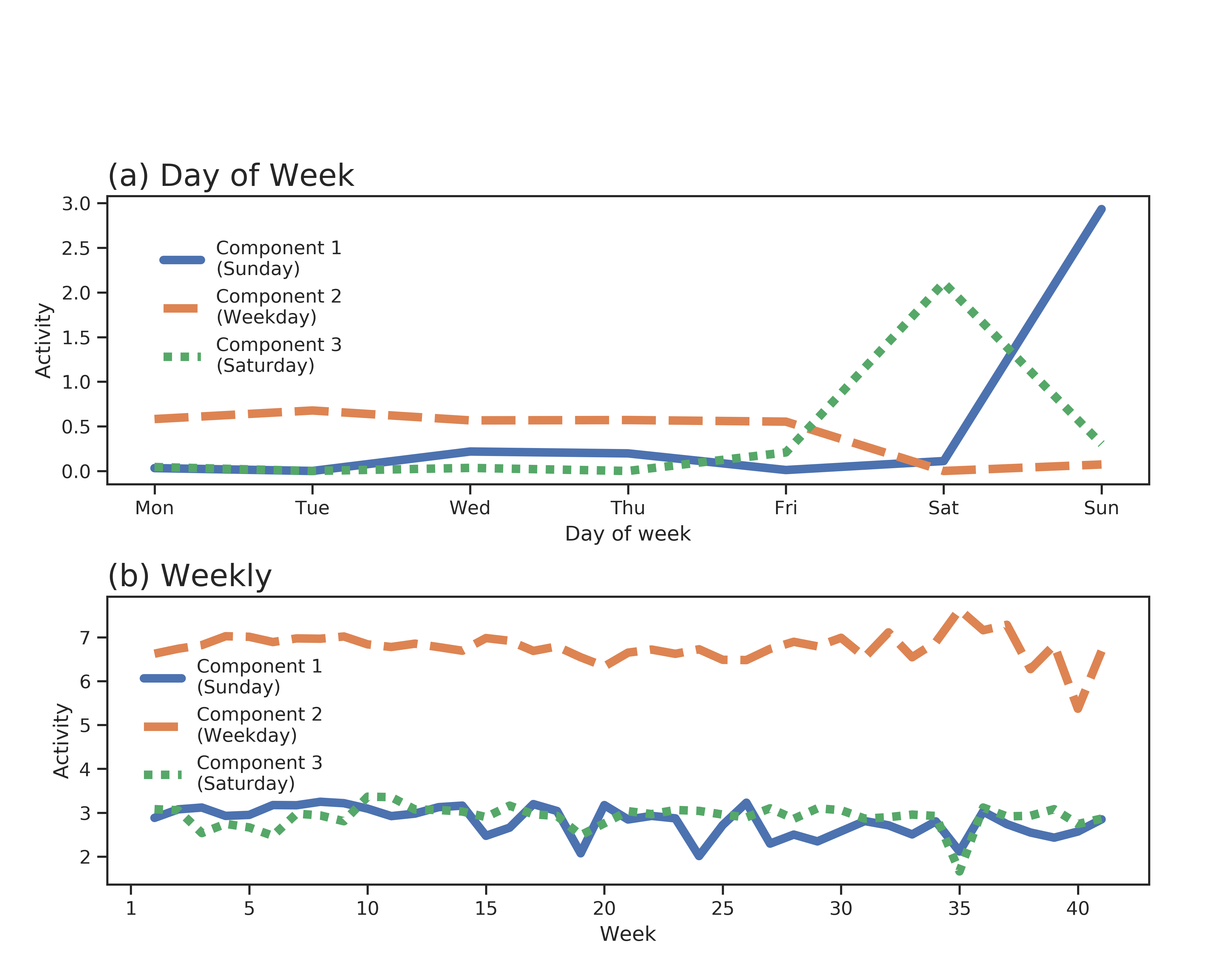}
  \caption{Activity at different timescales. (a) Day-of-week (i.e., intra-week) activity of each component. Activity of Component $r$ of day $j$ is given by $b_{jr}$. (b) Weekly (i.e., inter-week) activity of Component $r$ in week $k$ is given by $c_{kr}$.}
\label{fig:dow_activity}
\end{figure}

We firstly examine if the shopping activities have different dynamical patterns by looking at the components of day-of-week and weekly activities (\textbf{RQ1}). The $r$-th column of factor matrices $\mathbf{B}$ and $\mathbf{C}$ contain day-of-week and weekly activity patterns of Component $r$, respectively. For $R=3$, we find three distinctive day-of-week expenditure patterns from matrix $\mathbf{B}$ (Fig.~\ref{fig:dow_activity}a). Each pattern is characterized by the days of week on which activity is concentrated, namely \emph{Weekdays}, \emph{Saturday}, or \emph{Sunday}. This suggests that the users' expenditure behavior during a week is characterized by one of these three patterns or a combination of them.    

Similarly, weekly patterns can be extracted from $\mathbf{C}$ (Fig.~\ref{fig:dow_activity}b). Activity level of Component 2 (i.e., weekday-shopping pattern) is the highest among the three and relatively stable except for the last 5 weeks which correspond to the year end.
The activity of Component 1 (i.e., Sunday-shopping pattern) and 3 (i.e., Saturday-shopping pattern) are lower than that of Component 2 throughout the data period, while activity of Component 1 is a bit more volatile than that of Component 3.

\subsection{Expenditure patterns and demographic differences}

To address \textbf{RQ2}, we group the users based on their activities and see if each group has a characteristic demographic property. We use the factor matrix $\mathbf{A}$ obtained by the PARAFAC decomposition, on which we implement the $k$-medoids and the $k$-means methods to quantify the belongingness of user $i$ to each component. We compare the two clustering methods with silhouette analysis~\cite{kaufman2009finding} (Figs.~S1 and S2 in Supplementary Information (SI)).

We find that the $k$-medoids method gives us more evenly sized clusters compared to the $k$-means method (Figs.~S1 and S2). The mean silhouette coefficients for the $k$-medoids clustering are roughly the same across different numbers of clusters, which does not convey enough information to determine the number of clusters. We select the number of clusters $k=5$, judging from the fact that the rate at which the sum of distances between points in a cluster and the medoid decreases slows down around $k=5$ (Fig.~S3 in SI). In section~\ref{sec:characterize}, we will also show the results for which the consumers are grouped based on a threshold value.

Note that each consumer is classified by the $k$-medoids into one of the five non-overlapping groups based on their belongingness to each component quantified by matrix $\mathbf{A}$. To visualize the clustering result based on the $k$-medoids at the user level, we project the factor matrix $\mathbf{A}$ onto two-dimensional space by exploiting the t-SNE embedding~\cite{maaten2008visualizing} (Fig.~S4 in SI). The t-SNE is a visualization technique that allows us to convert high-dimensional data into low dimensional vectors~\cite{maaten2008visualizing}.

\subsection{Characterizing clusters based on the demographic properties} \label{sec:characterize}

Different multi-timescale expenditure patterns would reflect the users' demographic characteristics because the status of a consumer (i.e., age, gender, marital status, etc) might determine, at least partially, the timing of shopping and the variety of items purchased. Here, we compare the demographic characteristics among the five clusters identified by the $k$-medoids method.

\begin{figure}[tbh] 
\centering
 \includegraphics[width=.85\linewidth]{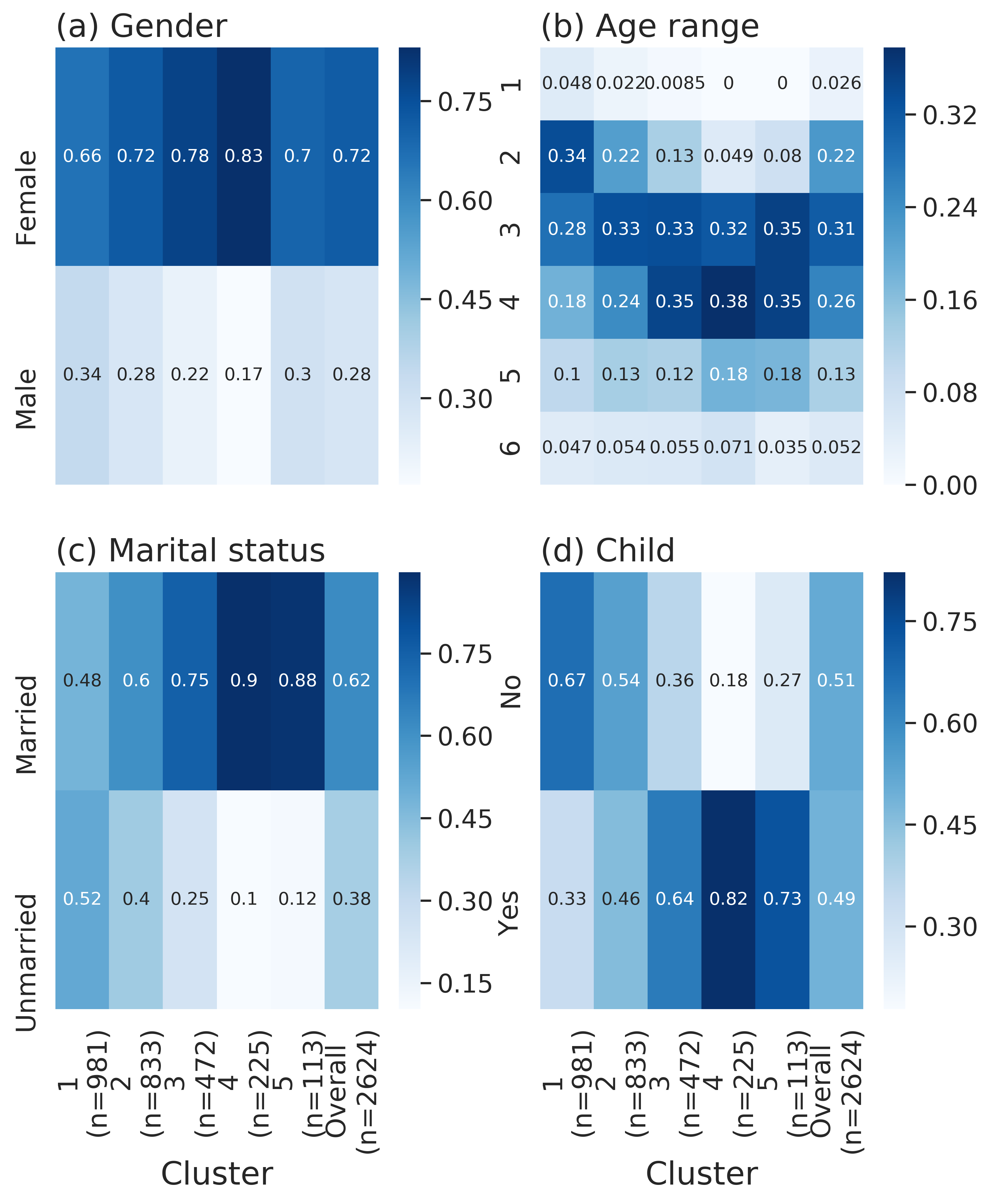}
  \caption{Demographic distribution for each cluster. (a) Gender, (b) Age range, from 1 (youngest) to 6 (oldest), (c) marital status, and (d) share of users who have or do not have children. }
  \label{fig:demo_cluster}
\end{figure}

 Fig.~\ref{fig:demo_cluster} indicates that each user cluster is characterized by some demographic properties. Typical examples can be found from Cluster 1 and Cluster 4. Cluster 1 consists of relatively young consumers having no children, while Cluster 4 appears to be formed mainly by married elderly women who have children. We use the chi-squared test to see if the demographic distribution in each cluster is significantly different from the null distribution obtained from the original demographic structure. The chi-squared statistic is given by the sum of squared differences between the number of users identified by the $k$-medoids method and the expected number under the null hypothesis: $\chi^{2}=\sum_{m}\sum_{\ell} \frac{(D_{\ell m}-E_{\ell m})^{2}}{E_{\ell m}}$, where $D_{\ell m}$ denotes the observed number of consumers in category $\ell$ (i.e., Male, Female, etc) for Cluster $m$, and $E_{\ell m}$ is the expected number of consumers in category $\ell$ for Cluster $m$ under the null~\cite{degroot2012probability}.

The results from the chi-squared tests suggest that for each demographic attribute (i.e., gender, age, marital status and child), the distribution of users identified by the clustering method is significantly different from the null distribution ($p < 0.001$). We also test whether there is a statistical difference in the distribution of users between two particular clusters. We conduct the statistical tests for all the pairwise combinations between different clusters.
For all the demographic attributes, the null hypothesis is rejected for most of the pairs of clusters (Table~S1 in SI).

\begin{figure}[tbh] 
\centering
 \includegraphics[width=.85\linewidth]{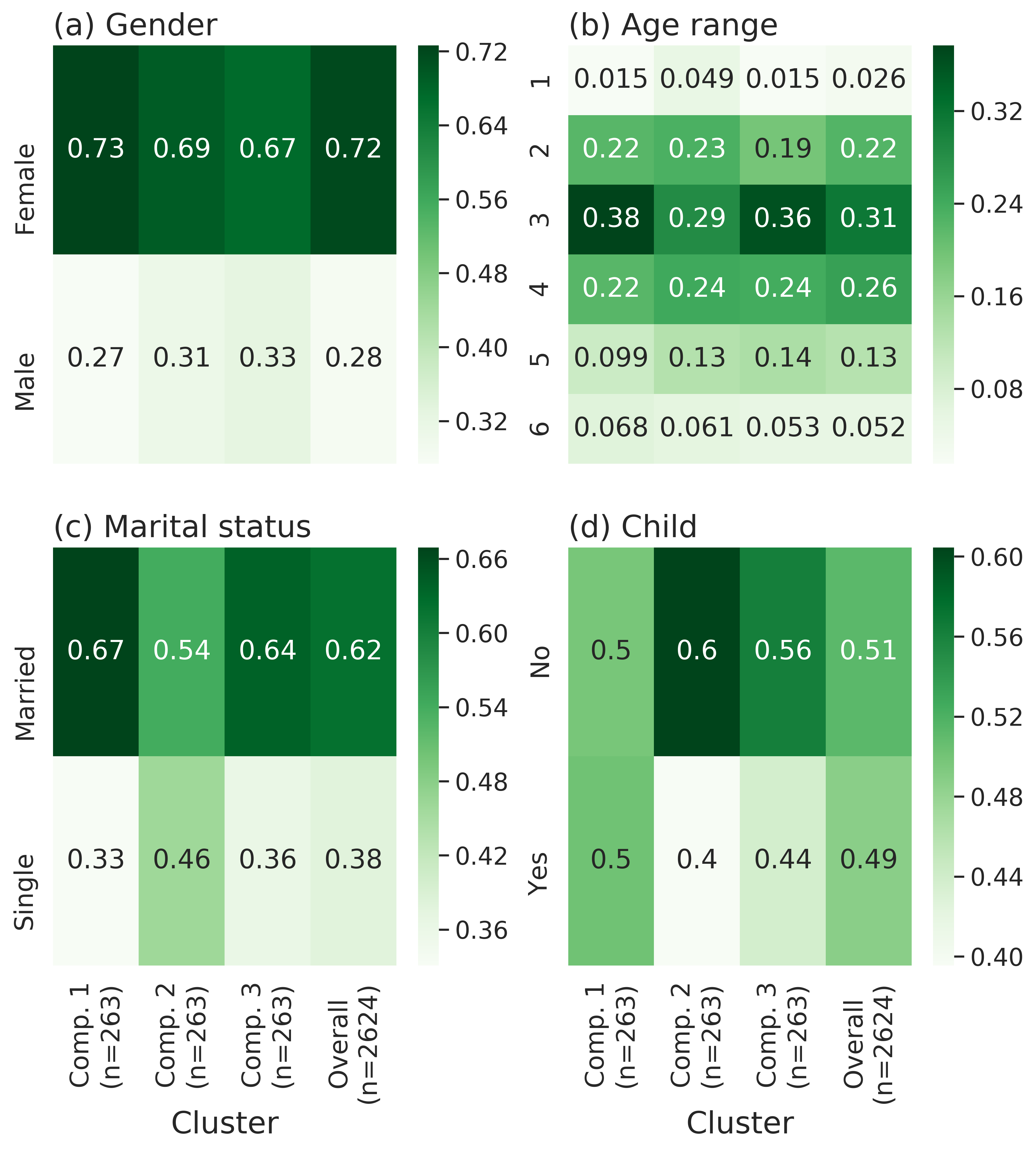}
  \caption{Demographic distribution of representative users in each component. User $i$ belongs to group $r$ if $a_{ir}/\sum_{r}a_{ir} \geq h_r$. (a) Gender, (b) Age range, from 1 (youngest) to 6 (oldest), (c) marital status, and (d) share of users who have or do not have a child.} 
  \label{fig:comp_demo}
\end{figure}

Lastly, we answer \textbf{RQ3} by focusing on representative users in each component, who are selected based on their belongingness to a component. Since the representative users in a given component would share similar demographic characteristics, we could identify which component is associated with which demographic properties.

We detect $R (=3)$ groups of representative users according to the following threshold rule: User $i$ is considered to belong to group $r$ if $a_{ir}/\sum_{r}a_{ir} \geq h_r$, where threshold $h_r$ is chosen such that only the upper 10 percent of users belong to group $r$.
Fig.~\ref{fig:comp_demo} shows the demographic distributions of the representative users belonging to each component. We note that each user may belong to multiple components, but such overlap is quite small (Fig.~S5 in SI).

We find that  ``Marital status" and ``Child" are two demographic properties that distinguish Component 2 (Weekday-shopping pattern) from the other components (Fig.~\ref{fig:comp_demo}c and d). For these two family-related attributes, the demographic distribution of the representative consumers in Component 2 is clearly different from the null distribution. This finding suggests that ``Marital status" and ``Child" would be the two driving factors that yield the five clusters detected by the $k$-medoids. On the other hand, the difference in user age between clusters seem to be more reflected in the activity of Component 1 (Sunday-shopping pattern) and 3 (Saturday-shopping pattern) rather than Component 2 (Fig.~\ref{fig:comp_demo}b), while it is not clear for gender (Fig.~\ref{fig:comp_demo}a). This means that gender and user age may be less important in extracting the multi-timescale patterns and the emergence of clusters classified by them.

\section*{Conclusion}

 We have presented a NTF-based method to extract dynamical shopping patterns of consumers from scanned receipt data collected through a bookkeeping application. The proposed method allows us to find intra- and inter-week expenditure patterns simultaneously, which would be impossible without such a large, high-resolution yet long time-series dataset. We found three multi-time scale patterns, each of which captures a characteristic expenditure behavior that is seen at daily and weekly scales. 

 While our method successfully revealed explicit patterns, there remain some issues that need to be addressed in future research. First, there may be other multi-timescale activity patterns that exist shorter and/or longer time scales rather than daily and weekly. For instance, the timing of shopping may be affected by time of a day, and consumption of expensive goods (e.g., cars) may be scheduled once in every ten years. Second, consumption patterns could also be encoded in what they purchased. While our analysis is based on the number of items purchased by a user, its composition would also be useful for revealing the demographic characteristics of users. Third, more multi-timescale patterns may exist in other economic and social contexts, such as financial markets, online communication networks and face-to-face networks. NTF is a useful and user-friendly tool for the detection of multi-timescale properties, and we hope our work will stimulate further research on many economic and social activities to better understand human behavior.  

\section*{Acknowledgements}
AM and EF are grateful to DARPA (grant no. D16AP00115). TK acknowledges financial support from JSPS KAKENHI Grant nos.~15H05729 and 19H01506. 

\section*{Conflicts of interest}
The authors have no conflicts of interest.

\newpage

\pagenumbering{arabic}
\setcounter{section}{0}
\setcounter{figure}{0}
\setcounter{table}{0}

\renewcommand{\thefigure}{S\arabic{figure}}
\renewcommand{\thesection}{S\arabic{section}}
\renewcommand{\thetable}{S\arabic{table}}

\begin{center}
{\Large{\textbf{Supplementary Information \\ \bigskip
``Detecting multi-timescale consumption patterns from receipt data: A non-negative tensor factorization approach"}}}\\ \bigskip
\large{Akira Matsui,
        Teruyoshi Kobayashi,
        Daisuke Moriwaki,
        Emilio Ferrara}

\end{center}

\begin{figure}[tbh] 
\centering
 \includegraphics[width=\linewidth]{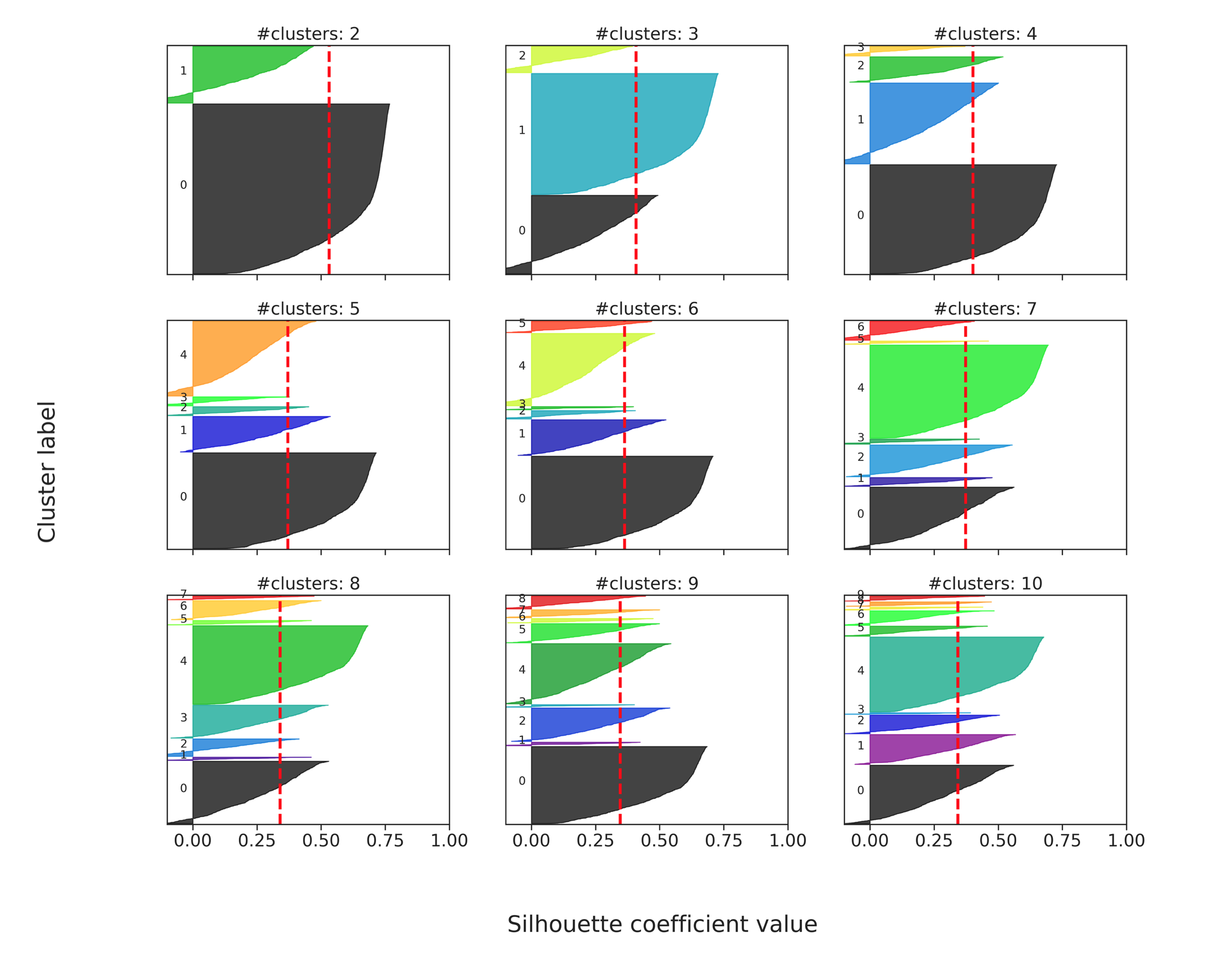}
  \caption{Silhouette analysis for the $k$-means clustering. Number of clusters is annotated at the top of each panel. Red dotted denotes the mean silhouette coefficient.} 
  \label{fig:shil_kmeans}
\end{figure}

\begin{figure}[tbh] 
\centering
 \includegraphics[width=\linewidth]{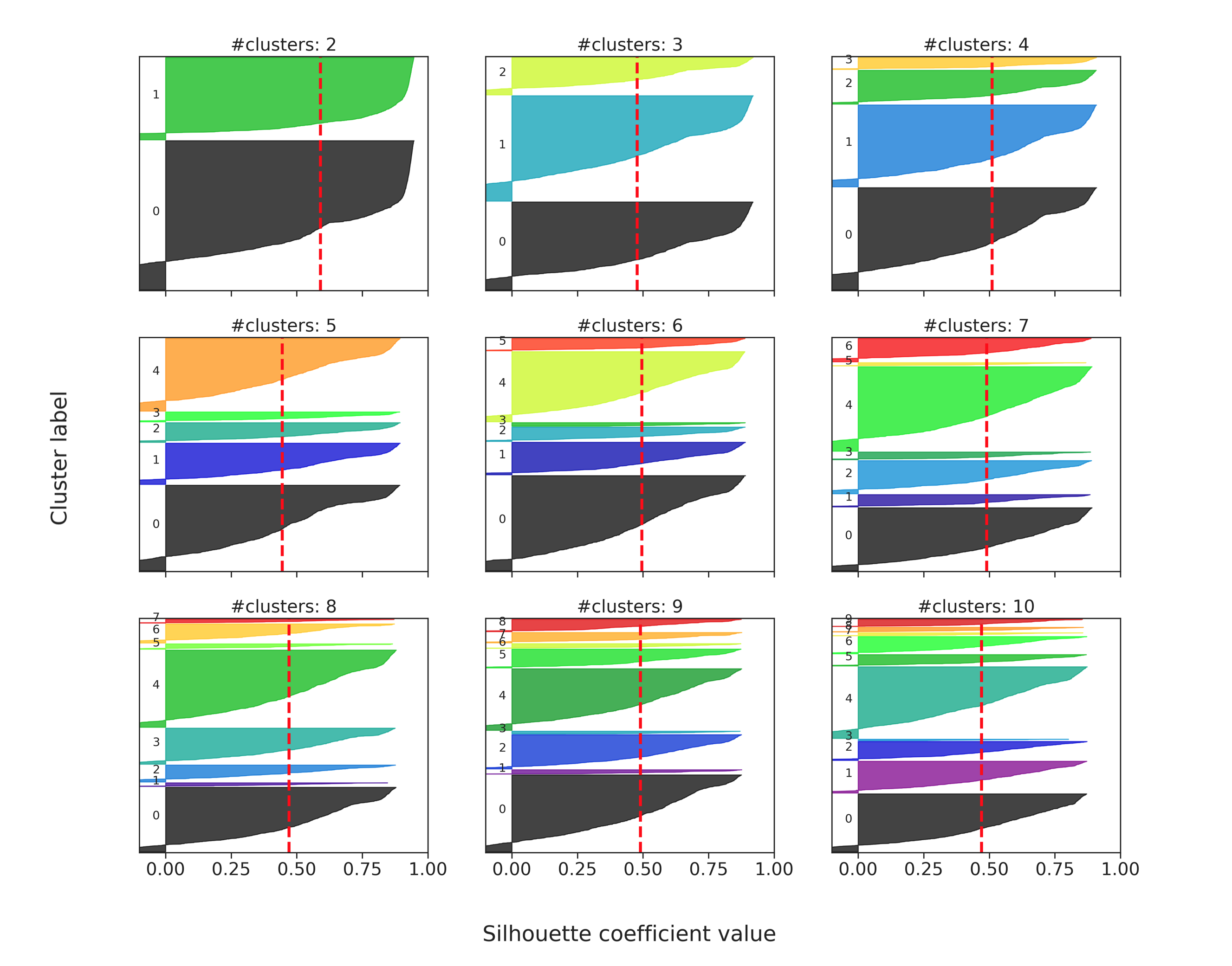}
  \caption{Silhouette analysis for the $k$-medoids clustering. Number of clusters is annotated at the top of each panel. Red dotted denotes the mean silhouette coefficient.} 
  \label{fig:shil_kmed}
\end{figure}

\begin{figure}[tbh] 
\centering
 \includegraphics[width=.8\linewidth]{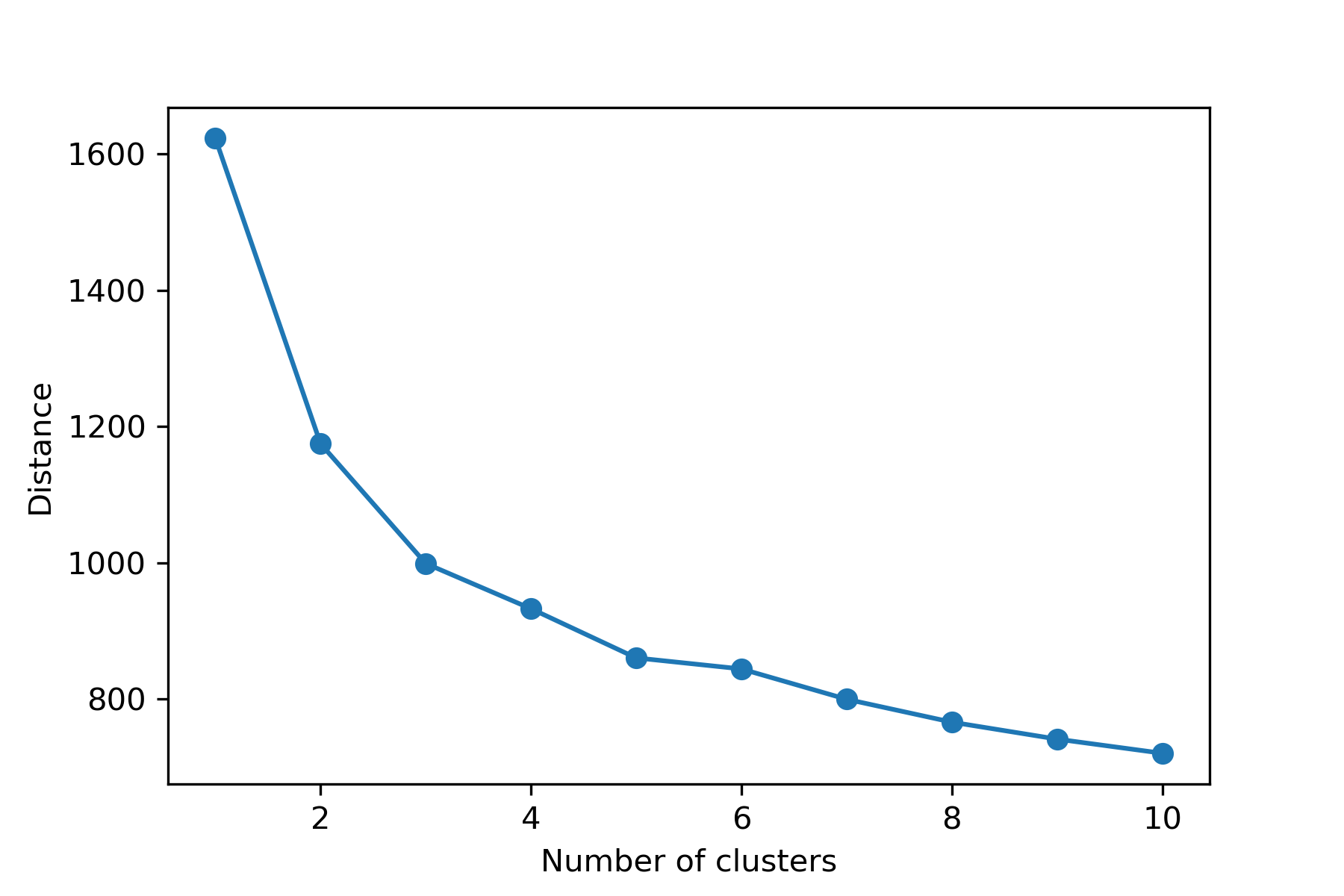}
  \caption{Sum of distances between points in a cluster and the medoid. We select $k=5$ for the analysis.} 
  \label{fig:elbo}
\end{figure}

\begin{figure}[thb]
\centering
 \includegraphics[width=10cm]{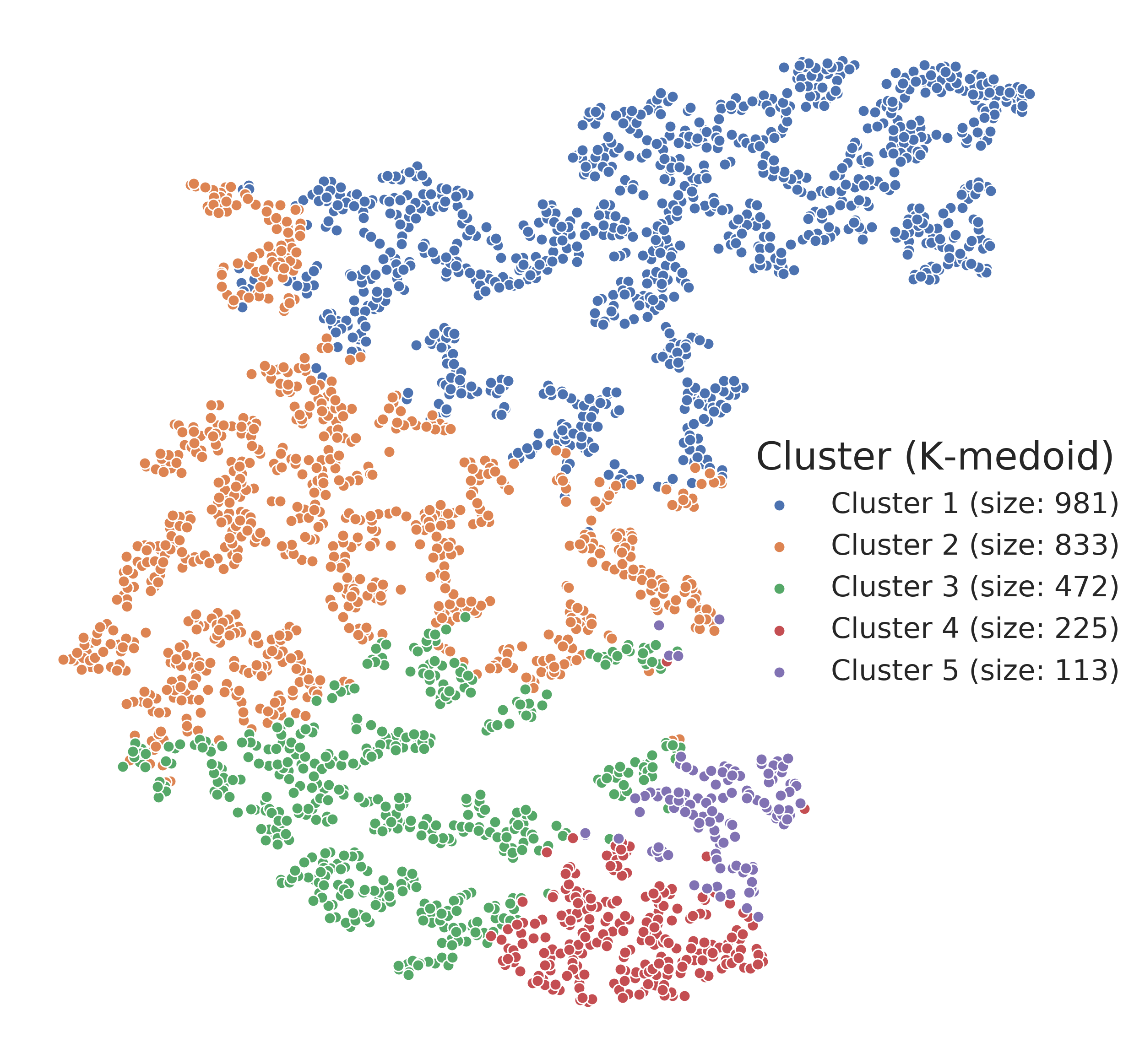}
  \caption{Clustering of users. The user feature vectors obtained from factor matrix $\mathbf{A}$ are visualized through the t-Distributed Stochastic Neighbor Embedding (t-SNE). } 
   \label{fig:tsne_with_kmedoid}
\end{figure}

\begin{figure}[tbh] 
\centering
 \includegraphics[width=.6\linewidth]{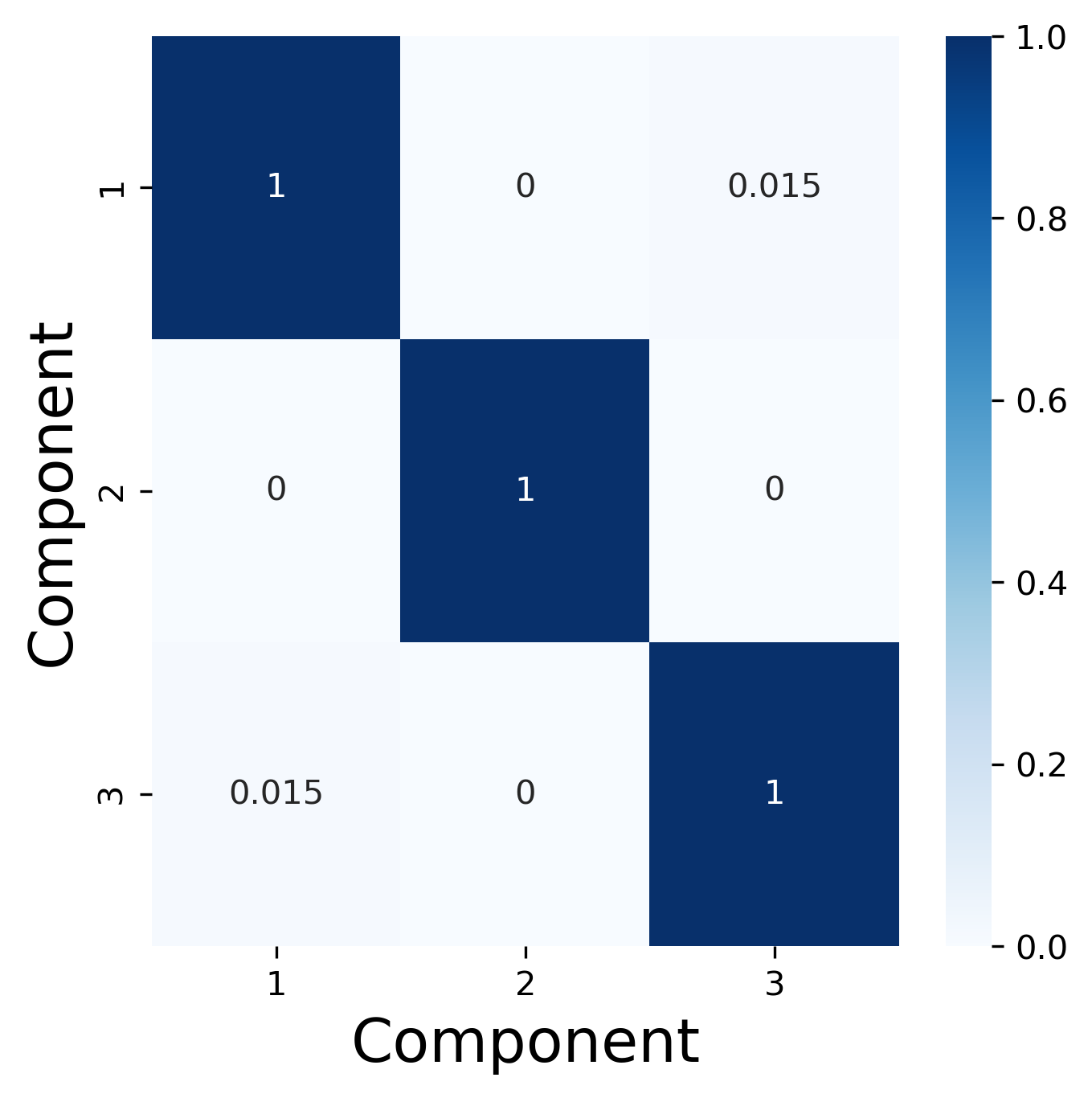}

  \caption{Jaccard index for the overlap of users belonging to multiple components.}
  \label{fig:comp_jaccard}
\end{figure}

\clearpage

\begin{table}[]
\centering
    \caption{Chi-squared test for demographic difference between clusters.} 
    \label{tab:chi_test}

\begin{tabular}{@{}lccrr@{}}
\toprule
          Attribute &  Cluster X &  Cluster Y &    $\chi^2$\hspace{.5cm} & Significance level \\
\midrule
Age range &          1 &          2 &    46.693 &        **** \\
     Age range &          1 &          3 &   105.398 &        **** \\
     Age range &          1 &          4 &   108.203 &        **** \\
     Age range &          1 &          5 &    48.561 &        **** \\
     Age range &          2 &          3 &    26.759 &        **** \\
     Age range &          2 &          4 &    47.835 &        **** \\
     Age range &          2 &          5 &    18.585 &          ** \\
     Age range &          3 &          4 &    16.049 &          ** \\
     Age range &          3 &          5 &     5.456 &             \\
     Age range &          4 &          5 &     3.213 &             \\
          Child &          1 &          2 &    31.310 &        **** \\
          Child &          1 &          3 &   121.161 &        **** \\
          Child &          1 &          4 &   179.896 &        **** \\
          Child &          1 &          5 &    70.022 &        **** \\
          Child &          2 &          3 &    37.733 &        **** \\
          Child &          2 &          4 &    92.999 &        **** \\
          Child &          2 &          5 &    29.783 &        **** \\
          Child &          3 &          4 &    24.574 &        **** \\
          Child &          3 &          5 &     3.788 &             \\
          Child &          4 &          5 &     3.524 &             \\
         Gender &          1 &          2 &     7.862 &          ** \\
         Gender &          1 &          3 &    22.765 &        **** \\
         Gender &          1 &          4 &    24.737 &        **** \\
         Gender &          1 &          5 &     0.642 &             \\
         Gender &          2 &          3 &     5.939 &           * \\
         Gender &          2 &          4 &    10.983 &         *** \\
         Gender &          2 &          5 &     0.274 &             \\
         Gender &          3 &          4 &     2.116 &             \\
         Gender &          3 &          5 &     3.673 &             \\
         Gender &          4 &          5 &     7.818 &          ** \\
 Martial status &          1 &          2 &    27.642 &        **** \\
 Martial status &          1 &          3 &    95.259 &        **** \\
 Martial status &          1 &          4 &   130.030 &        **** \\
 Martial status &          1 &          5 &    66.880 &        **** \\
 Martial status &          2 &          3 &    28.989 &        **** \\
 Martial status &          2 &          4 &    69.308 &        **** \\
 Martial status &          2 &          5 &    34.271 &        **** \\
 Martial status &          3 &          4 &    20.620 &        **** \\
 Martial status &          3 &          5 &     9.555 &         *** \\
 Martial status &          4 &          5 &     0.130 &             \\
\bottomrule

\end{tabular}
  \begin{tablenotes}
      \small
      \item $^{*}$p$<$0.1; $^{**}$p$<$0.05; $^{***}$p$<$0.01; $^{****}$p$<$0.001 (Bonferroni corrected).
    \end{tablenotes}
\end{table}

\end{document}